\documentclass[conference]{IEEEtran}

\usepackage{graphicx}
\graphicspath{{}{figures/}}
\usepackage{amsmath,amssymb,amsfonts}
\usepackage{booktabs,tabularx}
\usepackage{ragged2e}
\usepackage{algorithmic}
\usepackage{xcolor}
\usepackage{float}
\usepackage{placeins}

\usepackage[numbers]{natbib}
\usepackage[hidelinks]{hyperref}

\begin{document}

\title{Walking on Rough Terrain with Any Number of Legs}

\IEEEoverridecommandlockouts
\author{
Zhuoyang Chen,
Xinyuan Wang,
Shai Revzen
\thanks{The authors are with the University of Michigan, Ann Arbor, MI 48109, USA.
\{janchen, wxinyuan, shrevzen\}@umich.edu}
}

\maketitle

\begin{abstract}
Robotics would gain by replicating the remarkable agility of arthropods in navigating complex environments.
Here we consider the control of ``multi-legged'' systems which have 6 or more legs.
Current multi-legged control strategies in robots include large black-box machine learning models, Central Pattern Generator (CPG) networks, and open-loop feed-forward control with stability arising from the mechanics.
Here we present a multi-legged control architecture for rough terrain using a segmental robot with 3 actuators for every 2 legs, which we validated in simulation for 6 to 16 legs.
Segments have identical state machines, and each segment also receives input from the segment in front of it.
Our design bridges the gap between WalkNet-like event cascade controllers and CPG-based controllers: it tightly couples to the ground when present, but produces fictive locomotion when ground contact is missing.
It may be useful as an adaptive, computationally light-weight controller for multi-legged robots, and as baseline capability for scaffolding the learning of machine learning controllers.
\end{abstract}

\section{Introduction}

\subsection{Biological Inspiration: CPGs and WalkNet}

Animals exhibit remarkable abilities to move efficiently through complex and unpredictable environments. 
Cockroaches and stick insects are classic examples of robust hexapod locomotion on rough terrain \citep{Holmes2006}. 
These observations have inspired decades of research in biomechanics, neurophysiology, and bio-inspired robotics. 
Early physiological studies on central pattern generators (CPGs) established that vertebrates and many invertebrates possess neural circuits capable of producing rhythmic activity without requiring step-by-step sensory input \citep{Wilson1961, Marder2001, Hultborn2006, Edgerton1976, Bidaye2018, Pearson1972}, forming the core mechanism underlying walking, swimming, and flying.

Complementary to CPG-based explanations, the Walknet model provides a biologically grounded, decentralized account of six-legged locomotion. 
Walknet shows that stable insect gaits can emerge without strong central oscillators, relying instead on distributed sensory-driven coordination \citep{Cruse1998, Schilling2013}. 
Through local rules and the exploitation of body-environment coupling, such as positive velocity feedback, Walknet naturally reproduces tripod and tetrapod gait patterns, adapts to uneven terrain, and remains robust to disturbances such as leg overload, loss of ground contact, or even leg removal. 
In this sense, Walknet supplements CPG theories by illustrating that rhythmic stability can arise purely from sensory-driven, distributed control \citep{Schilling2013}. 
At the same time, the model is compatible with the possibility that central oscillators may take over in conditions where sensory input is compromised, allowing for the possibility that insects flexibly combine decentralized reflexive control and central pattern generation depending on environmental demands \citep{neveln2019information}.

\subsection{Hexapod Locomotion: RHex and Related Platforms}

For multi-legged robots, stability can arise mechanically, allowing for feed-forward control strategies, and robustness in the face of diverse and unpredictable environments. 

Several groups explored hexapedal robots, exploiting their natural stability and exploring insect-inspired ideas. 
Early examples such as \textbf{Robot III} and \textbf{Biobot} \citep{Nelson1997} directly modeled the cockroach leg.
The \textbf{LAURON I--V} series drew inspiration from the stick insect \citep{LauronV2013}.
Whereas \textbf{RiSE} \citep{Autumn2005, Haynes2006} wove together ideas from gecko attachment to surfaces, the use of tails for pitch control, and the advantages of 6 legs to produce an exceptional climbing robot that could also walk.

Among the many bio-inspired hexapods, \textbf{RHex} stands out as one of the most compelling and often replicated examples \citep{Saranli2001, Altendorfer2001}.
Renowned for its speed and robustness, RHex can traverse fractured, obstacle-dense, and irregular terrain without sensing or adaptation.
Additional work further explored RHex's dynamics, such as self-righting \citep{Saranli2004}, stair descent \citep{Campbell2003}, and later transformable variants \citep{RHexT3_2021}. 

Despite its success, RHex also exhibits clear limitations: having 1-DoF per leg leaves it with little freedom for selecting footholds or adapting its motion to the shape of the terrain \citep{Campbell2003, Altendorfer2001}. 
The lack of rich sensory feedback further limits its performance in perception-dependent tasks.

\subsection{CPG-Based Control Methods and Learning Approaches}

In motor control research, bio-inspired ``\textit{Central Pattern Generators (CPGs)}'' have been broadly used across robots that swim, walk, and slither \citep{Ijspeert2007, Crespi2008, Rutishauser2008}. 
Comprehensive reviews such as \citep{Ijspeert2008} and representative works \citep{Arena2004, Inagaki2003, Klaassen2002} demonstrate the effectiveness of CPGs in generating stable rhythmic locomotion.
CPG, used as a control method, are typically modeled as a system of coupled oscillators, with each joint assigned an oscillator. 
Through appropriate coupling, the system produces stable limit-cycle behaviors that can rapidly recover from perturbations, enabling robust gait generation.
Researchers have demonstrated open-loop CPG systems achieving fast locomotion even on rough terrain \citep{Sprowitz2011Oncilla, Rutishauser2008}. 
Later studies incorporated sensory feedback to realize closed-loop CPG control \citep{Righetti2008}.
Sartoretti et al.\ proposed an approach that adapts CPG parameters using inertial sensory feedback \citep{Sartoretti2018}. 
By leveraging body pose, orientation, and height information, the limit cycles of coupled oscillators can adjust themselves in joint space to enable locomotion and body posture control.

Various oscillator models---including Hodgkin-Huxley, Kuramoto, Hopf and van der Pol oscillators \citep{Arena2004, Inagaki2003, Righetti2008, Veskos2005}---have been implemented for CPG control. While effective, CPG-based approaches generally lack systematic design principles: coupling topology, phase biases, and gains must often be tuned manually or optimized via heuristics\citep{Ijspeert2008}.

Recently some researchers have proposed learning-based CPG optimization methods \citep{Bellegarda2022, Zhang2024}, and even methods blending CPG structures with modern deep reinforcement learning \citep{Deshpande2023, Liu2020}. 
The lead to the breakthrough result of a fully learned locomotion controller demonstrated on the \textbf{ANYmal} robot \citep{Lee2020}, highlighting the potential of learning-driven locomotion, albeit at the cost of heavy computation and large-scale data requirements.

\subsection{State-Machine Control}

While complex, compute- and sensing-intensive control approaches have produced amazing results, other than the work on RHex, we know little about what simple control and low complexity can accomplish.
Considering how easily CPG-based control dynamics adapt to alternative oscillator types, we reconsidered whether continuous should be used at all.
We abstracted the essential functionality of a CPG into a cycle of four discrete, task-relevant states and implement them through a finite state machine (FSM). 
This formulation preserves features of rhythmic control, such as phase progression and modularity, while avoiding the complexity of oscillator dynamics, extensive parameter tuning, or high training costs. 
We then coupled a segmental state-machine oscillator producing yaw to two leg oscillators raising and lowering the legs.
Finally, we allowed each segmental oscillator to influence the oscillator directly behind it.
The resulting controller remained simple in structure, but it is capable of generating coordinated and adaptive swing--stance behaviors on irregular and unpredictable terrain.

On the mechanical side, we addressed some of the limitations of RHex's leg design with a minimalistic extension. 
To strike a balance between ``structural simplicity'' and ``sufficient locomotive capability,'' we introduced a design that has 3 actuators for every pair of legs, allowing each foot to move within a hemispherical region, allowing foot placement to be adapted to the terrain features.

Finally, we added a WalkNet inspired sensory ability -- the ability to detect whether a leg has contacted the ground and is bearing load.

\section{Method}

\subsection{Hardware Design}

\begin{figure*}[htbp]
    \centering
    \includegraphics[width=0.78\textwidth]{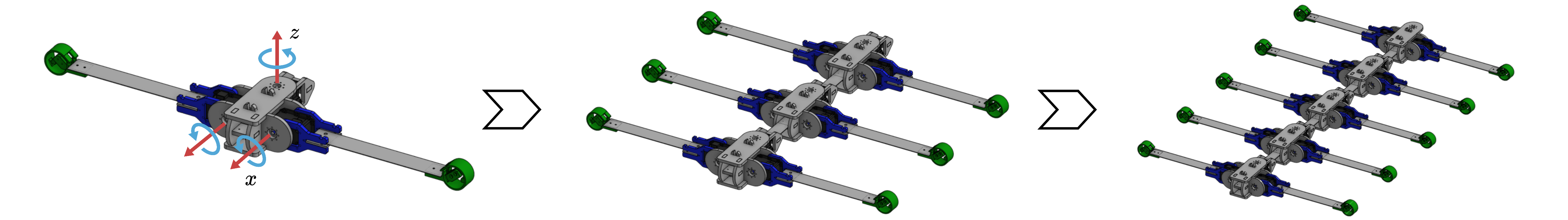}
    \includegraphics[width=0.18\textwidth]{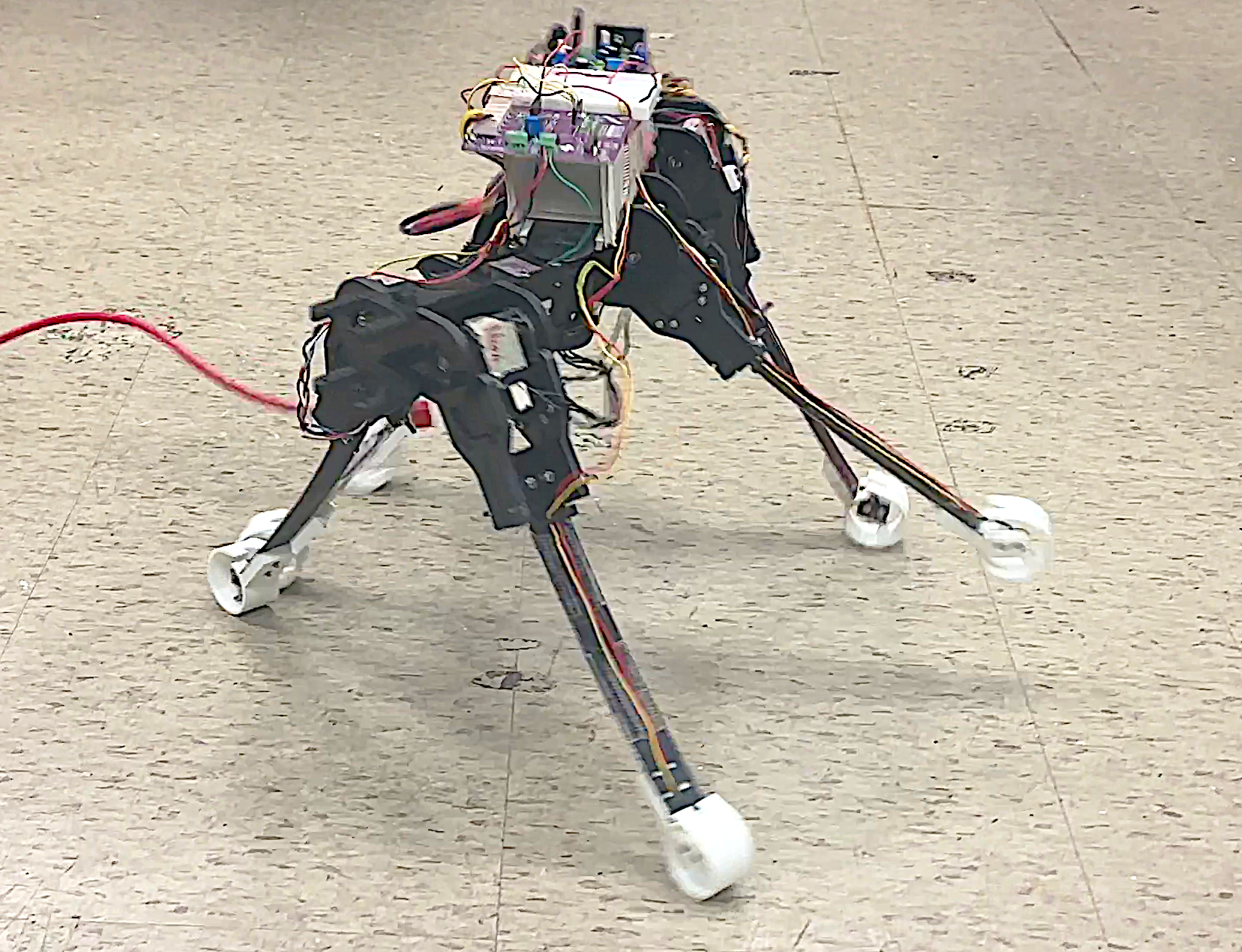}
        \caption{Modular mechanical design of the robot. %
        A segment (left) has 3 DOF: two leg motors rotate about the x-axis, and a yaw motor rotates about the z-axis. %
        This modular design can be chained with backbone joints to create a hexapod (middle), or, e.g. a decapod (right).
        We build a hexapedal version (right, photo)}
    \label{fig:modular}
\end{figure*}

We designed the hexapedal version of our robots (Fig.~\ref{fig:modular}) as an 8-DoF system inspired by prior ``centipede''-based designs~\citep{Sastra2012, Sastra2008}.
The prior ``centipede'' robots relied on two spine actuators and one leg actuator for every pair of contra-lateral legs.
The advantage of the ``centipede'' design was that it allowed for non-slip locomotion since its planar projection was a 4-bar linkage between each pair of stance legs (see \citep{zhao2020multi} for detailed analysis). 
However, in the ``centipede'' one actuator moved both legs on the same segment, which prevented independent height control of legs. 
Instead of the Z-X-Z (yaw-roll-yaw) configuration of the centipede, we used a Z-(XX) configuration: a single backbone yaw motor, followed by a two roll-axis motors, one to the right and one to the left. 
This configuration decoupled bilateral leg actuation allowing us to independently control the heights of the right and left foot, at the expense of making slipping inevitable.
Prior work suggested that not only is slipping not problematic for multi-legged robots, it is actually advantageous \citep{zhao2020multi}.

Following the bio-inspired approach in~\citep{Sastra2012, Sastra2008}, we employed compliant legs whose stiffness we chose to match the stiffness appearing in the Spring-Loaded Inverted Pendulum (SLIP) template~\citep{Cavagna1977, Blickhan1993} of running animals \citep{Dickinson2000}. 
We chose the cantilever properties such that (1) standing on 3 legs, each at roll angle $45^\circ$, the robot would sink $10\%$ of its unloaded height; and (2) in this position, the aspect ratio length to width was approximately that of cockroaches.

We designed all structural components with simple geometries suitable for laser cutting and engraving, eschewing slow and fracture prone 3D printing alternatives. 
We assembled the body segments primarily through geometric mating constraints rather than fasteners; we used screws only for motor mounting and compliant-joint connections, and ensured that all the screws were of the same family (M3 socket head). 
By chaining additional backbone joints, we could extend the platform from a hexapod to an octopod or even longer configurations (Fig.~\ref{fig:modular}). 

The physical robot we built (Fig.~\ref{fig:modular} rightmost) had the segments rigidly linked to each other.
However, to allow for longer robots on rough terrain in the simulation, we designed a backbone bending joint which is compliant in pitch, but rigid in yaw and resistant to twisting. 
Such a backbone element is challenging to build in practice because of the unusually high torsional stiffness required, and its design and fabrication will be the topic of a future publication.
These backbone bending elements provided an elastic, passive degree of freedom in the pitch direction while avoiding a freedom in roll which would have neutralized much of ability of the legs to lift off.

We performed all CAD work using Autodesk Inventor and Onshape; CAD files will be made open-source after publication.


\subsection{Locomotion Control}

Our controllers were finite state machines. 
In each segment, they comprised a segmental ``yaw oscillator'' with 4 states, including two composite states containing 3 sequential sub-states, respectively, controlling the yaw motor, and ``right leg'' and ``left leg'' 4-state oscillators, each controlling its respective leg.

\begin{figure*}[htbp]
    \centering
    \includegraphics[width=0.8\linewidth]{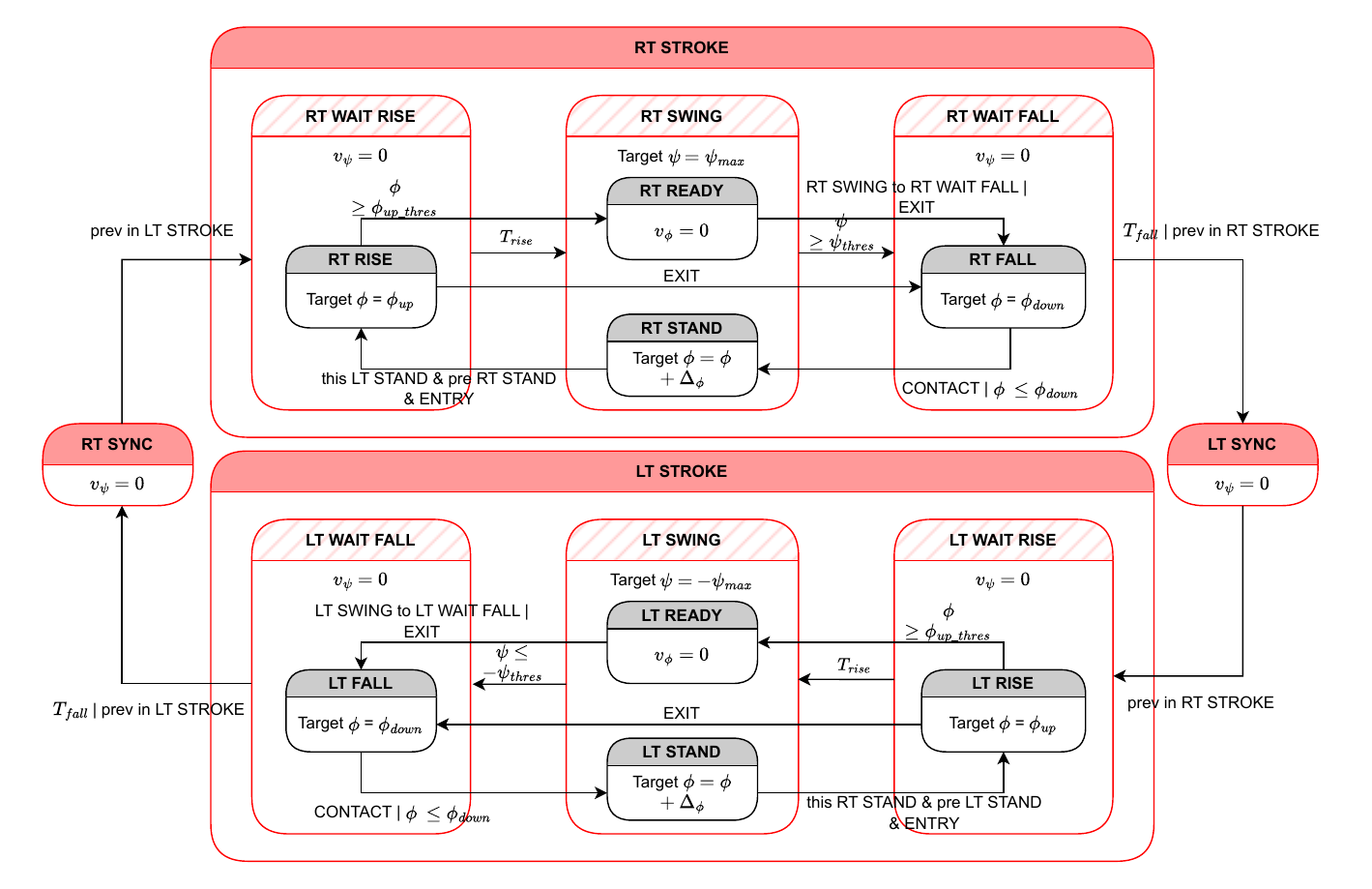}
    \caption{State machine architecture used for hierarchical yaw-leg coordination.}
    \label{fig:fsm}
\end{figure*}

Figure~\ref{fig:fsm} illustrates the structure of the controller state-machine, including all transitions. 
The red states represent the yaw-joint FSM, while the gray states represent the left and right leg FSMs. 
We denoted the leg roll angles as $\phi$ and the yaw angle as $\psi$. 
We defined positive leg swing as upward, and positive yaw rotation as clockwise when viewed from above. 

We coupled the yaw controllers unidirectionally from head to tail through a \texttt{SYNC} state that blocked execution until the preceding segment entered the composite \texttt{STROKE} state, which is divided into three sequential sub-states \{\texttt{WAIT RISE}, \texttt{SWING}, \texttt{WAIT FALL}\}. 
This enforced a stable phase offset between adjacent segments thereby creating a coordinated traveling wave along the body.

Each yaw FSM also hierarchically modulated its two leg FSMs. 
When the yaw FSM entered the \texttt{STROKE} state, it activated the leg FSMs to initiate lift-off. 
The yaw FSM temporarily remained in \texttt{WAIT RISE} sub-state to allow each leg FSM to reach the \texttt{READY} state once its roll angle $\phi$ exceeded a threshold. 
If the yaw FSM exited \texttt{STROKE} prematurely, the leg FSM transitioned immediately to \texttt{FALL} to avoid deadlock. 
During the transition from sub-state \texttt{SWING} to sub-state \texttt{WAIT FALL}, the yaw FSM triggered the leg to complete foot touchdown. 
Upon contact detection or $\phi$ falling below the lower limit, the leg returned to \texttt{STAND} and updated its desired position.

All actuators used PD control with an internal trapezoidal velocity profile generator. 
The velocity profile generator attempted to generate the velocity trajectory with best effort, utilizing the maximum speed $v_{max}$ during the acceleration and deceleration phases.
We discovered that if the legs stop their roll motion when ground contact is detected, the height of the body ratchets down over multiple cycles, and the robot crashes into the ground.
Therefore, every time the foot reached the ground, we added a touchdown offset so that the feet extended downward a little further.

The locomotion control architecture induced several control inputs or parameters listed in Table~\ref{tab:parameters}. 
Although the mechanical design allowed the motor to have $-\pi/2$ to $\pi/2$ freedom, our experiments indicated that a roll angle larger than $-1.0$~rad yielded the most stable walking performance. 
Similarly, we constrained the yaw motor range within $[-0.6, 0.6]$ to prevent the feet from interfering.




\begin{table}[t]
\renewcommand{\arraystretch}{0.9}
\footnotesize
\centering
\caption{Control Parameters}
\label{tab:parameters}
\begin{tabularx}{\columnwidth}{@{} l >{\RaggedRight\arraybackslash}X l l l @{}}
\toprule
Parameter & Description & Range & Value & Unit \\ 
\midrule
$\psi_{max}$ & Max yaw angle & $[-0.6,0.6]$ & 0.6 & rad \\

$\psi_{thres}$ & Yaw threshold (SWING→WAIT FALL) 
& $(-\psi_{max},\psi_{max})$ & 0.55 & rad \\

$\phi_{up}$ & Max leg roll 
& $[-1.0,0]$ & -0.1 & rad \\

$\phi_{up\_thres}$ & Roll threshold (RISE→READY) 
& $[-1.0,\phi_{up})$ & -0.15 & rad \\

$\phi_{down}$ & Min leg roll 
& $[-1.0,\phi_{up\_thres})$ & -1.0 & rad \\

$\Delta\phi$ & Touchdown offset 
& $[0,\phi_{down})$ & 0.6 & rad \\

$T_{rise}$ & WAIT RISE delay 
& $\ge 0$ & 0.1 & s \\

$T_{fall}$ & WAIT FALL delay 
& $\ge 0$ & 0.1 & s \\

\bottomrule
\end{tabularx}
\end{table}

By adjusting the parameters above, we tuned the behavior of the finite state machine controller. 
For instance, changing the maximum yaw angle $(\psi_{max})$ modulated the step size of each leg, affecting the walking speed; adjusting $(\phi_{up})$, $(\phi_{down})$, and $(\Delta \phi)$ resulted in different body heights and foot clearances, affecting the robot's ability to cross obstacles. 
When the robot walked on extremely rough terrain (with large variance in the vertical direction), we also increased $T_{rise}$ and $T_{fall}$ to give the leg enough time to rise and fall before swinging, thereby avoiding tripping or tipping over, with a small sacrifice in walking speed. 
Changing the motor's maximum speed had a similar, more intuitive effect on walking speed, and we did not detail those effects here.

\subsection{Simulation Setup}

\begin{figure}[H]
    \centering
    \includegraphics[width=0.6\linewidth]{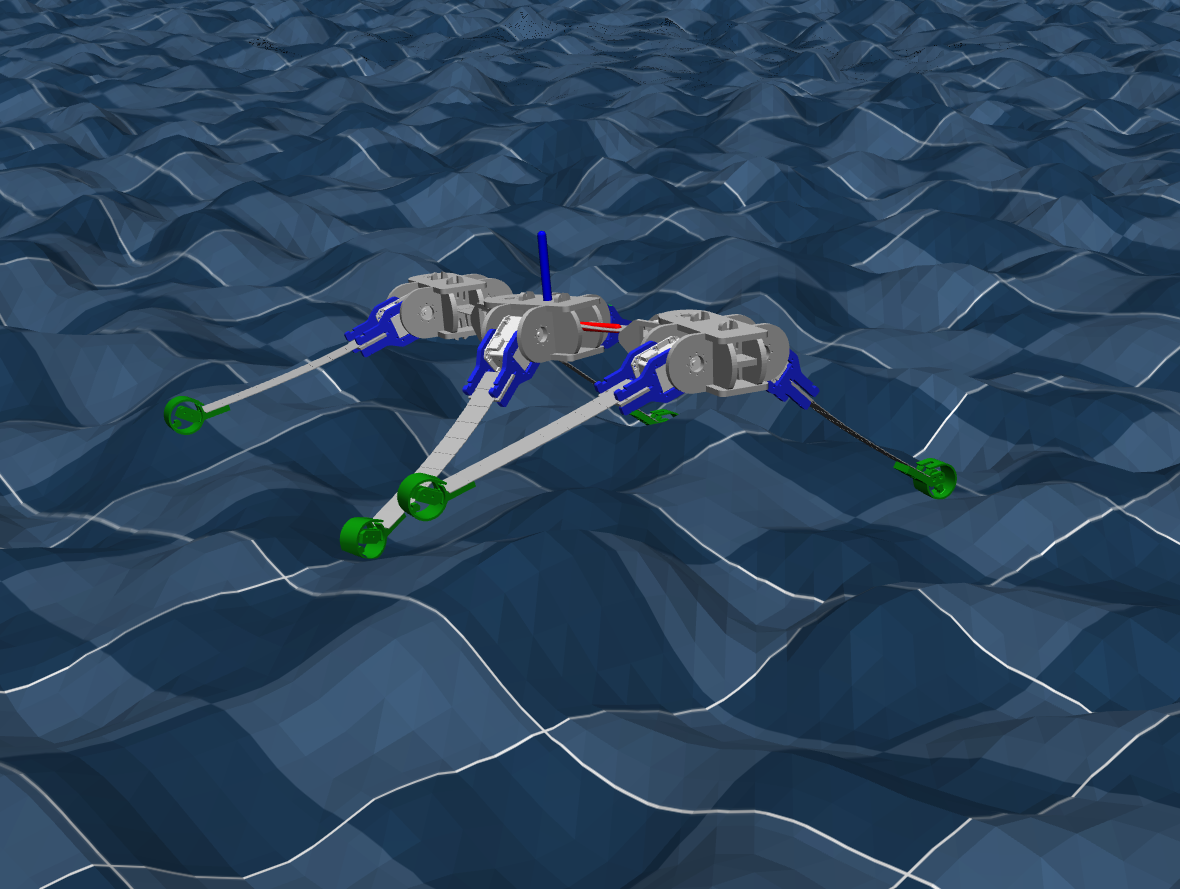}
        \caption{The 3-segment hexapod robot simulated on a rough terrain in the MuJoCo simulator.}
    \label{fig:sim-demo}
\end{figure}

We tested the robot model and locomotion control, and collected the results using the MuJoCo simulator (version 3.3.5; Fig.~\ref{fig:sim-demo}). 
We controlled the simulated robot using a \texttt{pygame} based library used for many internal projects in our lab.

We converted the robot CAD model to a compatible robot description file using the onshape-to-robot package and assigned the mass properties for each part in Onshape as follows.

\begin{table}[htbp]
\centering
\caption{Simulated robot properties}
\label{tab:properties}

\vspace{2mm} 

\begin{tabular}{@{}lll@{}}
\toprule
Part & Property & Value \\ \midrule
All 3D-printed / laser-cut parts & PLA density& 1.24 g/{cm}$^3$ \\
Compliant parts & 300 stainless steel density & 8.0 g/{cm}$^3$ \\
Dynamixel MX-64 & Mass & 126 g \\
~ & No-load speed & 78 RPM \\
~ & Holding torque & 7.3~N$\cdot$m \\ \bottomrule
\end{tabular}
\end{table}

We simulated the compliant parts by dividing the part into  three equal segments connected by hinge joints, with torsional stiffness chosen to produce the overall bending as specified above.
 
We set the materials and motor characteristics to match the  \textbf{Dynamixel MX-64} motors\footnote{from \url{https://emanual.robotis.com/docs/en/dxl/mx/mx-64/}} and actual materials used in the physical robot (see Table \ref{tab:properties}).
We provided the parameters defining the FSM controller in the simulation in Table \ref{tab:parameters}.

\section{Simulation Results}
In this section, we evaluated the robot's walking stability and control robustness. 
We defined stability through two primary lenses: (1) Synchronization, determining whether the distributed control architecture yielded a stable, rhythmic gait by synchronizing all segments under varying initial conditions; and (2) Terrain Robustness, determining whether the generated gait maintained body posture and progression across various irregular terrains.

\subsection{Phase Estimation and Coordinate Definition}
Since we coupled the state machines via transition rules and they operated cyclically, they functioned as coupled oscillators, each possessing a local period and phase that contributed to a global gait cycle. 
To quantify synchronization, we analyzed whether the state machines achieved a phase-locked state.
However, because the state machines contained static states (e.g., the \texttt{STAND} state of legs or \texttt{SYNC} state of yaw motors), traditional linear phase estimation methods like the Hilbert Transform were unsuitable. 
Instead, we utilized the phase estimation method developed in~\citep{Revzen2008-es}. 
Specifically, we constructed a 7-dimensional signal for each segment--comprising the $x$ and $z$ coordinates of the foot tips, the yaw motor angle, and leg roll angles--to estimate the instantaneous phase of each segment. 
We calculated the phase difference as the circular difference between the circular average phases of even-numbered and odd-numbered segments: $\arg\left((\sum_{k\in\text{even}} \phi_k) / (\sum_{k\in\text{odd}} \phi_k)\right)$ where the $\phi_k$ are phasors.

To evaluate physical stability, we analyzed the robot's body height, pitch, and roll. 
As the robot is a multi-segmented articulated body without a fixed torso, we defined a virtual body frame $O_G$:
\begin{enumerate}
    \item Origin ($O_G$): We defined this as the geometric centroid of all segment positions, $O_{G} = \frac{1}{N}\sum^{N}_{i=0} s_i$, where ${s_i}$ was the world-frame position of the $i_{th}$ segment.
    
    \item Forward Axis ($\hat{X_{G}}$): We computed this via Principal Component Analysis (PCA). 
    We calculated the covariance matrix $C = \sum^{N}_{i=0}(s_i-O_{G})(s_i-O_G)^T$ and selected the eigenvector associated with the largest eigenvalue as $\hat{X_{G}}$.
    
    \item Vertical Axis ($\hat{Z_{G}}$): We derived this from the mean orientation of the segments. 
    Let $\hat{z}_i$ be the local $z$-axis of the $i_{th}$ segment. 
    We computed the mean vector $\mathbf{\bar{z}} = \frac{1}{N}\sum^{N}_{i=0} \hat{z}_i$ and orthonormalized it against $\hat{X_{G}}$ using Gram--Schmidt projection: $\mathbf{v} = \mathbf{\bar{z}} - (\mathbf{\bar{z}} \cdot \hat{X_{G}})\hat{X_{G}}$; $\hat{Z_{G}} = \frac{\mathbf{v}}{\|\mathbf{v}\|}$.
    
    \item Lateral Axis ($\hat{Y_{G}}$): We completed the frame via the right-hand rule: $\hat{Y_{G}} = \hat{Z_{G}} \times \hat{X_{G}}$.
\end{enumerate}

\subsection{Experimental Setup}
We simulated two robot configurations -- a 3-segment (6-leg) hexapod and an 8-segment (16-leg) hexadecapod -- across 20 trials with randomized initial phase configurations. 
To validate the controller's adaptability, we tested five distinct environments:
\begin{itemize}
    \item Floating: We set gravity to 0 with no ground contact, isolating the intrinsic controller dynamics in a set-up reminiscent of ``fictive locomotion'' \cite{grillner1979central}.
    
    \item Flat: We used an infinite planar surface.
    
    \item Rough: We generated terrain using a Gaussian Process (z-scale $= 0.3$, frequency $\approx$ segment length).
    
    \item Hill: We used quadratic slopes ranging from $-13.3^\circ$ to $13.3^{\circ}$, ensuring the Center of Mass (COM) remained within the support polygon.
    
    \item Stairs: We constructed 10 upward and 10 downward steps (length $0.2$~m, height $0.04$~m), approximating the segment length and half the foot clearance, respectively.
\end{itemize}

\subsection{Synchronization}
\begin{figure}[htbp]
    \begin{minipage}{0.5\textwidth}
        \centering
        \includegraphics[width=\textwidth,trim={0 1cm 0 3cm}, clip]{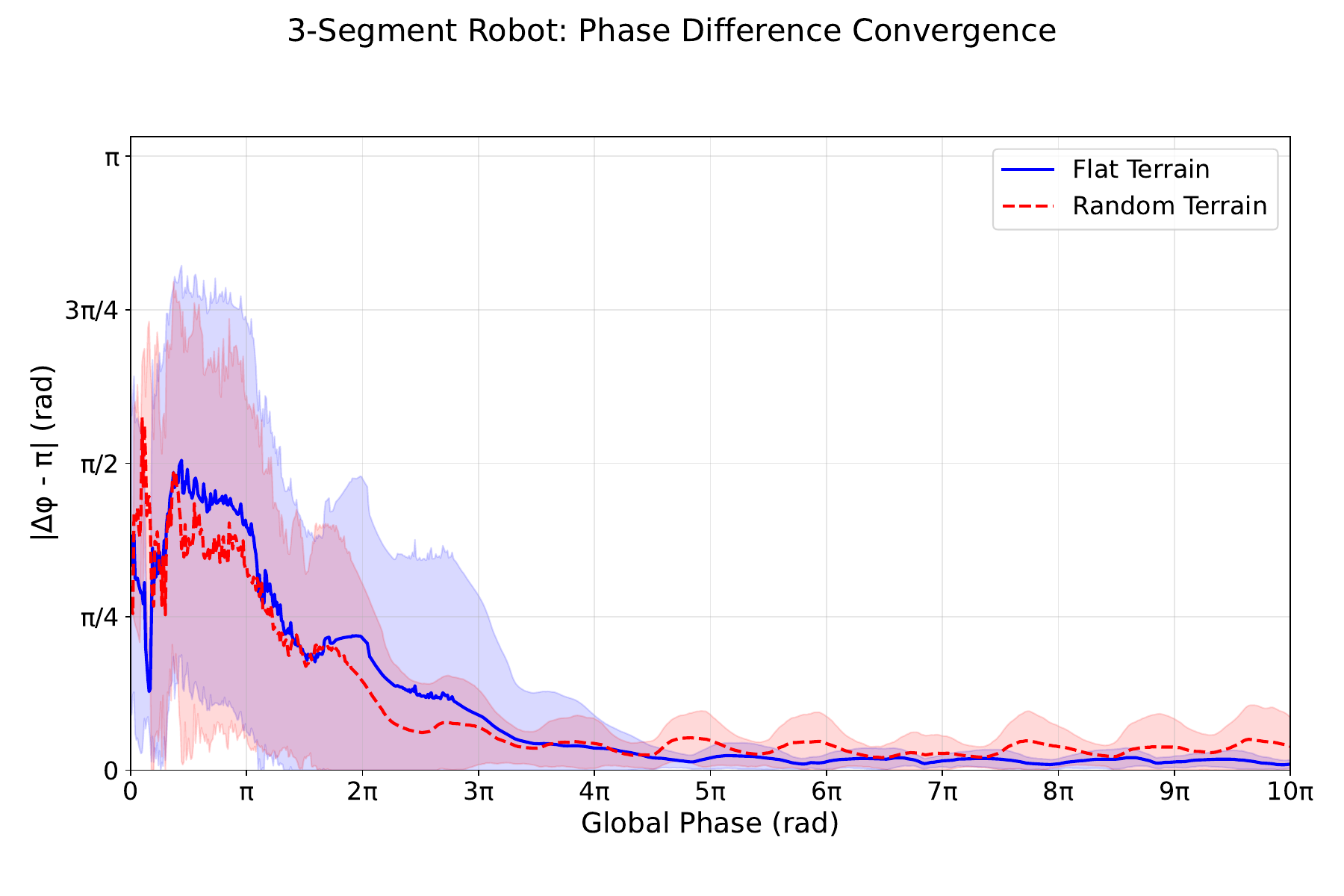} 
        {3-segment hexapod robot \href{https://youtu.be/-wFcwE2lJVg}{[video]}}
        \label{fig:transient-3-seg}        
    \end{minipage}
    \begin{minipage}{0.5\textwidth}
        \centering
        \includegraphics[width=\textwidth,trim={0 1cm 0 3cm}, clip]{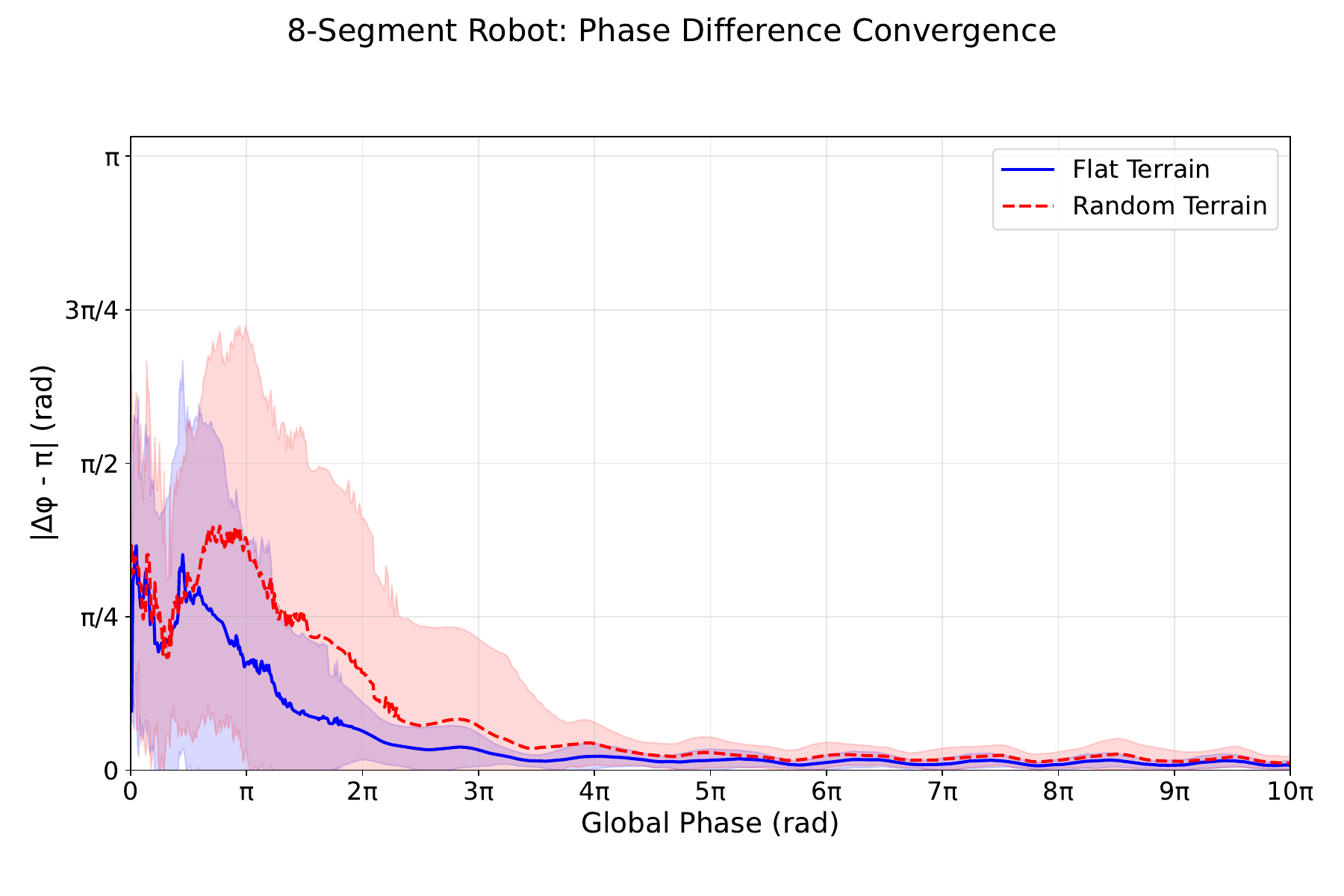} 
        {8-segment hexadecapod robot \href{https://youtu.be/IaNMZ3O1xCo}{[video]}}
        \label{fig:transient-8-seg}
    \end{minipage}
    
    \caption{Robustness of Phase Convergence. The phase error decays rapidly on both ideal Flat Terrain (Blue) and Random Rough Terrain (Red Dashed). Despite the significant ground irregularities in the random terrain case, the distributed controller successfully synchronizes the legs, demonstrating that the gait stability is robust to environmental noise. This performance is consistent across both the (a) 6-legged and (b) 16-legged configurations.}
    \label{fig:transient-combined}
\end{figure}

A critical requirement for distributed controllers is the ability to recover a coordinated gait from arbitrary initial conditions, even in the presence of external disturbances. 
Figure \ref{fig:transient-combined} illustrates the transient response of the phase error (the deviation from the ideal alternating tripod gait) for both robot configurations.

On Flat Terrain (blue traces), the phase error exhibited a smooth exponential decay, converging to zero within 2 periods. 
This indicated that the local interactions between segments rapidly drove the system to the stable alternating tripod limit cycle.

Crucially, the system maintained this performance on the random terrain (red dashed traces). 
Despite irregular foot contacts and variable terrain heights disrupting the individual leg cycles, the collective phase error still converged. 
While the steady-state error on rough terrain showed slightly higher variance due to more segments, the controller successfully enforced the global gait pattern. 
This demonstrated high robustness: the distributed FSM did not require a perfectly flat surface to synchronize; rather, the synchronization was compliant enough to absorb terrain irregularities.

Furthermore, this behavior scaled effectively. Figure \ref{fig:transient-8-seg} shows that the 8-segment (16-leg) robot achieved synchronization on a similar timescale to the 3-segment version. 
The local signaling rules were sufficient to globally stabilize the gait, despite more than doubling the number of segments.

\subsection{Terrain Robustness}
\begin{figure*}[htbp]
    \centering
    \includegraphics[width=0.9\linewidth, trim={3cm 0cm 0cm 0cm}, clip]{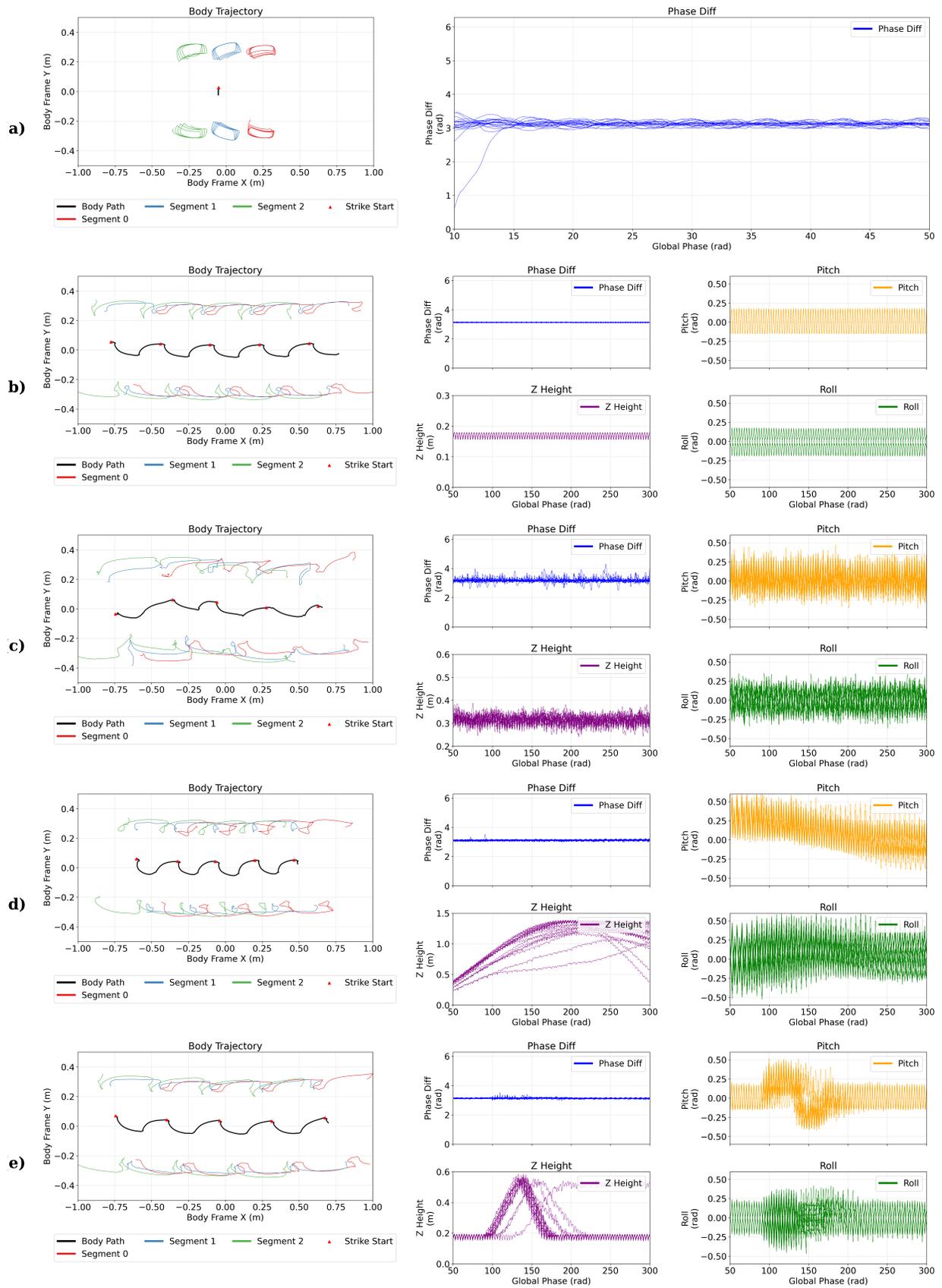}
    \caption{Body and foot trajectories, phase difference, and body posture convergence for a 3-segment robot simulated on (a) floating terrain, (b) flat terrain, (c) random terrain, (d) hill terrain, and (e) stair terrain. We omitted the body posture in the floating scene as it is not well defined}
    \label{fig:combined-plot-3seg}
\end{figure*}

\begin{figure*}[htbp]
    \centering
    \includegraphics[width=0.9\linewidth, trim={3cm 0cm 0cm 0cm}, clip]{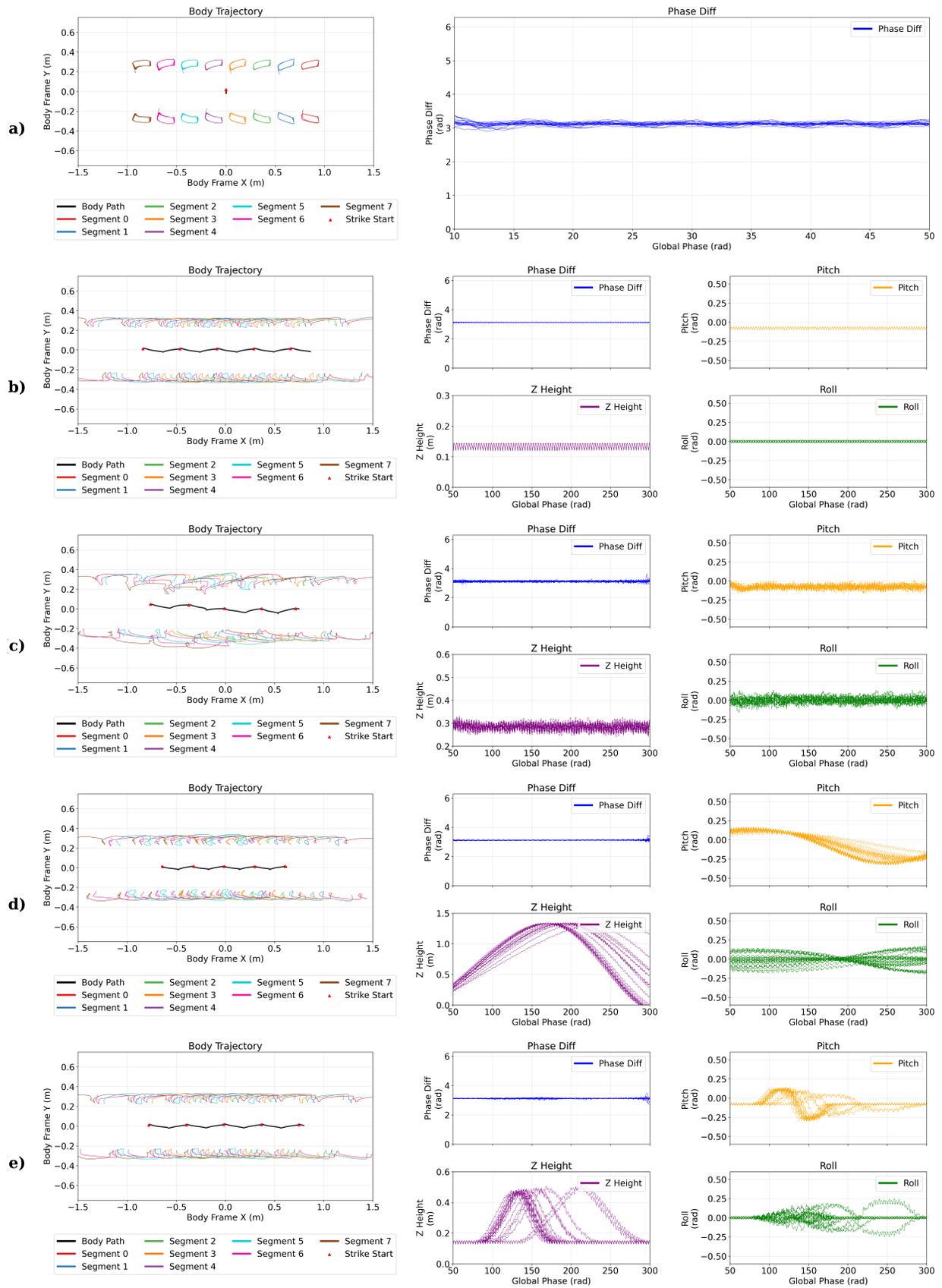}
    \caption{Body and foot trajectories, phase difference, and body posture convergence for an 8-segment robot simulated on (a) floating terrain, (b) flat terrain, (c) random terrain, (d) hill terrain, and (e) stair terrain. We omitted the body posture in the floating scene as it is not well defined}
    \label{fig:combined-plot-8seg}
\end{figure*}

We visualized the steady-state performance across terrains in Fig.~\ref{fig:combined-plot-3seg} and Fig.~\ref{fig:combined-plot-8seg}. 
In the Floating condition, the controller produced a stable ``fictive locomotion'' limit cycle. 
With no ground contact to trigger transitions, the system defaulted to a rhythmic pattern where the even-odd phase difference converged to a constant offset, which revealed the intrinsic oscillator dynamics of the FSM network.

On Flat terrain, the pattern anchored to the physical environment, producing steady forward progression with a nearly straight body path. 
Pitch, roll, and body height settled into small, bounded oscillations. 
When the robot traversed the Hill terrain, it maintained the anti-phase traveling wave while adapting its pitch and height to the slope. 

On Rough and Stair terrains, the benefits of the distributed architecture were most evident. 
While steps and local irregularities sharply deformed individual foot trajectories (visible as kinks in the foot traces), the global phase pattern remained robust. 
The body height tracked the terrain profile closely, and the body posture remained stable. 
Comparing the 3-segment and 8-segment trials, we observed qualitatively identical behavior; the 16-legged robot traversed the same obstacles with equal stability, which further validated that the proposed FSM acted as a modular, scalable pattern generator capable of navigating diverse environments.

\section{Discussion}

The simulations presented in this paper suggested that a remarkably simple, segment-wise finite state machine could generate robust multi-legged locomotion over a wide range of terrains. 
With a single set of parameters, the controller produced a consistent traveling-wave pattern that phase-locked reliably from randomized initial conditions for both a 3-segment hexapod and an 8-segment hexadecapod. 
The resulting gaits were not only stable in the sense of convergent phase relationships, but also maintained reasonably bounded body pitch, roll, and height while traversing flat ground, smooth hills, random rough terrain, and stairs. 
In the floating condition, the same controller exhibited fictive locomotion, which indicated that the FSM network itself embodied an intrinsic limit cycle rather than relying on carefully tuned continuous oscillators. 
Together, these outcomes supported the view that discrete, event-triggered pattern generators could capture much of the functionality typically associated with CPG-based designs, while remaining easy to implement and scale.
Crucially, unlike many dynamic models, our FSM was governed by a small number of easy to interpret parameters.

At the same time, our simulation study made several assumptions that limit how directly the results could be transferred to hardware.
One such assumption was embedded in the bending element attaching segments to each other.
We assumed it can only allow bend in the pitch direction, with no twisting around the X axis. 
While such torsionally stiff bending joints can be fabricated, their design is non-trivial.
A leaf spring such as the CAD model would suggest has sufficient torsional compliance to significantly hamper the ability of the legs to push against contra-lateral legs in the segments before and after, leading to dramatically reduced leg clearance. 

Another limitation arose our choice to switch from Z-X-Z ``centipede'' segments of prior work to the Z-(XX) design used here. 
The Z-X-Z design seemed to have a highly desirable property -- its projection on the horizontal plane, under the assumption the feet are pin joints, is a four-bar linkage which has a kinematic degree of freedom. 
This DoF implies that the Z-X-Z design can do a grounded walk without slipping, whereas our Z-(XX) design cannot, even on flat terrain.
As expected, the simulated feet exhibited slipping which was particularly noticeable on flat and gently sloped terrain. 

Humans and other large land mammals are highly averse to slipping, and we tend to project this as a design requirement for our robots.
It has been shown that (1) allowing slipping can dramatically improve performance \cite{zhao2020multi}; (2) multi-legged animals such as cockroaches are constantly slipping as they move \cite{sachdeva2018cockroaches}; and (3) the presence of dynamic coulomb friction slipping does not alter the principally kinematic nature of the motion \cite{zhao2022walking}. 
Our results here showed that despite all the slipping, the robot still progressed forward reliably across all tested terrains with the current body plan and controller, and the systematic presence of slipping did not adversely affect the ability to move.
From a design perspective this suggests that for multi-legged locomotion which is passively stable accepting some slip in exchange for a reducing the weight an complexity from $2$~motors/leg to $1.5$~motors/leg may be worthwhile. 

There were also more conventional places the model was unrealistic. 
We modeled friction as isotropic and time-invariant, whereas real surfaces exhibit spatial variation, wear, and contamination. 
The rough terrain was still a rigid and smooth 2.5 dimensional topography, unlike real gravel and soil which flow and crumble, and have overhangs.
The controller itself did not incorporate explicit foothold planning, perception, or online adaptation; it reacted only through local contact information and the simple touchdown-offset rule. 

\subsection{Future Work}
Looking forward, several directions appeared promising. 

On the mechanical side, installing a backbone flexure with high torsional stiffness in the actual robot and putting realistic values for this flexure in the simulation is a critical step in bringing these design ideas more fully into fruition.
Designing and testing alternative foot structures which have anisotropic friction and some axial compliance could reduce slipping while preserving low motor per leg ratio \citep{hatton2025finsler}.
Finally, replacing the contact-switch based ground contact sensing with a strain gauge on the legs could make ground contact detection much more robust to mechanical damage and e.g. puddles or other terrain features that would damage distal electronics.

On the control side, the FSM presented here could serve as a scaffold for more sophisticated behaviors.
For example, higher-level modules could modulate state transition thresholds or desired joint ranges based on sensory feedback, or learning-based policies could be layered on top to adjust parameters online while preserving the underlying phase-locked structure. 

We could also incorporate additional feedback. An IMU might help with detecting large pitch and roll excursions, and perhaps allow the robot to handle larger slopes by feedback those data.
A magnetometer or other absolute heading sensor might allow the robot to maintain its overall heading on very rough terrain.

Finally, a systematic comparison with continuous CPG-based controllers and with more reactive, WalkNet-style schemes on the same hardware would help clarify the relative strengths and weaknesses of discrete pattern generators like the one proposed here. 

Overall, while the present work was limited by idealized models and simplified mechanics, it demonstrated that a modular FSM controller combined with appropriately designed compliant structures could provide a robust and scalable bio-inspired foundation for legged locomotion over challenging terrain.

\bibliography{references}
\end{document}